%% file: main.tex
\documentclass[10pt,twocolumn,letterpaper]{article}

\usepackage{iccv}              

\usepackage{bbding}

\input{macros}
\input{math_commands}

\definecolor{iccvblue}{rgb}{0.21,0.49,0.74}
\usepackage[pagebackref,breaklinks,colorlinks,allcolors=iccvblue]{hyperref}

\usepackage{booktabs}
\usepackage{xcolor}
\usepackage{lipsum}
\usepackage{amssymb}
\usepackage{multirow}
\usepackage[<options>]{animate}
\usepackage{graphicx}
\usepackage[symbol]{footmisc}
\usepackage{pifont}
\usepackage{makecell}
\usepackage{algorithm}
\usepackage[noEnd,indLines]{algpseudocodex}

\usepackage[skins,theorems]{tcolorbox}

\title{Tool-R1: Sample-Efficient Reinforcement Learning for Agentic Tool Use}

\author{Yabo Zhang\textsuperscript{1} \ \ \
Yihan Zeng\textsuperscript{2}\ \ \ 
Qingyun Li\textsuperscript{1}\ \ \
Zhen Hu\textsuperscript{1}\ \ \ 
Kavin Han\textsuperscript{1}\ \ \ 
Wangmeng Zuo\textsuperscript{1(\Envelope)}\\
\textsuperscript{1}Harbin Institute of Technology \quad
\textsuperscript{2}Huawei Noah’s Ark Lab \\
}

\begin{document}
\maketitle

\input{text/0_abstract}
\input{text/1_introduction}
\input{text/2_related_work}
\input{text/3_method}
\input{text/4_experiment}

{
    \small
    \bibliographystyle{ieeenat_fullname}
    \bibliography{main}
}
\clearpage
\input{text/appendix}
\end{document}

%% file: macros.tex
\newcommand{\ignorethis}[1]{}

\makeatletter
\DeclareRobustCommand\onedot{\futurelet\@let@token\@onedot}
\def\@onedot{\ifx\@let@token.\else.\null\fi\xspace}

\def\eg{\emph{e.g}\onedot} 
\def\ie{\emph{i.e}\onedot}

\makeatother

\newrobustcmd{\B}{\bfseries}
\newcommand*{\rom}[1]{\expandafter\romannumeral #1}

\definecolor{mydarkblue}{rgb}{0,0.08,1}
\definecolor{mydarkgreen}{rgb}{0.02,0.6,0.02}
\definecolor{mydarkred}{rgb}{0.8,0.02,0.02}
\definecolor{mydarkorange}{rgb}{0.40,0.2,0.02}
\definecolor{mypurple}{RGB}{111,0,255}
\definecolor{myred}{rgb}{1.0,0.0,0.0}
\definecolor{mygold}{rgb}{0.75,0.6,0.12}
\definecolor{myblue}{rgb}{0,0.2,0.8}
\definecolor{mydarkgray}{rgb}{0.66,0.66,0.66}

\newcommand{\myparagraph}[1]{\vspace{-6pt}\paragraph{#1}}


%% file: math_commands.tex
\def\eqref#1{equation~\ref{#1}}

\def\1{\bm{1}}



%% file: text/0_abstract.tex
\begin{abstract}
Large language models (LLMs) have demonstrated strong capabilities in language understanding and reasoning, yet they remain limited when tackling real-world tasks that require up-to-date knowledge, precise operations, or specialized tool use. To address this, we propose Tool-R1, a reinforcement learning framework that enables LLMs to perform general, compositional, and multi-step tool use by generating executable Python code. Tool-R1 supports integration of user-defined tools and standard libraries, with variable sharing across steps to construct coherent workflows. An outcome-based reward function, combining LLM-based answer judgment and code execution success, guides policy optimization. To improve training efficiency, we maintain a dynamic sample queue to cache and reuse high-quality trajectories, reducing the overhead of costly online sampling. Experiments on the GAIA benchmark show that Tool-R1 substantially improves both accuracy and robustness, achieving about 10\% gain over strong baselines, with larger improvements on complex multi-step tasks. These results highlight the potential of Tool-R1 for enabling reliable and efficient tool-augmented reasoning in real-world applications.
Our code will be available at \url{https://github.com/YBYBZhang/Tool-R1}.
\end{abstract}

%% file: text/1_introduction.tex
\input{figText/main_demo}
\section{Introduction}
Large language models (LLMs)~\cite{instructgpt,guo2025deepseek,shao2024deepseekmath,jaech2024openai,qwen2.5} have made significant progress in natural language understanding and generation, especially in challenging tasks such as mathematical reasoning and code generation. These strengths stem from the internal knowledge acquired during large-scale pretraining. However, LLMs still perform poorly on real-world problems that require up-to-date facts or specialized expertise. For example, they may struggle to query the latest weather conditions or accurately analyze massive structured data. To address these limitations, recent works~\cite{yao2023react,shen2023hugginggpt,lu2025octotools,yang2023mm,suris2023vipergpt,qin2023toolllm} have focused on equipping LLMs with external tools such as search engines and code interpreters, so they can access accurate information and perform precise operations beyond their built-in knowledge.

Existing methods enable LLMs to use external tools during the reasoning process through prompt engineering~\cite{yao2023react,lu2025octotools,suris2023vipergpt,qian2023creator} or reinforcement learning (RL)~\cite{jiang2025deepretrieval,jin2025search,zheng2025deepresearcher}, but both approaches face fundamental limitations in handling dynamic real-world scenarios. 
Prompt engineering methods rely entirely on the internal knowledge to determine tool usage, making them inherently unable to adapt to the diverse and unpredictable feedback from real-world environments. 
This leads to poor robustness when encountering unexpected responses or environmental changes. 
While RL-based methods demonstrate superior capability in learning from various types of environmental feedback, they face two critical scalability challenges. 
First, current methods typically rely on JSON-formatted tool calls that are restricted to predefined user-provided tools or APIs, limiting their ability to leverage diverse tool combinations and create custom tool configurations necessary for complex real-world tasks. 
Second, the prohibitive cost of tool execution during training further confines existing RL approaches to specific domains (\eg, retrieval task~\cite{jin2025search}) and prevents broader applicability.

In this work, we introduce Tool-R1, a novel method that supports general and compositional tool use across complex real-world tasks, with LLMs trained through sample-efficient reinforcement learning. 
Unlike prior RL-based methods using JSON-formatted tool calls, Tool-R1 first adopts a more flexible and general tool-use mechanism by generating executable Python code. 
This design not only supports user-defined tools (\eg, web browser and multimedia parser) and standard Python libraries, but also enables intermediate variable sharing across steps, allowing LLMs to construct coherent and multi-round tool-use workflows.
During fine-tuning, we employ an outcome-based reward function to encourage LLMs to freely explore when and how to use appropriate tools: an LLM judges whether the predicted answers align with open-ended reference answers, while the success rate of code execution serves as an auxiliary reward to promote reliability and correctness.
However, training such tool-use policies through reinforcement learning can be time- and resource-intensive, as it often involves frequent online sampling and tool execution during fine-tuning. To alleviate this burden, Tool-R1 maintains a dynamic sample queue that caches recent high-quality trajectories, which are then reused to improve sample efficiency and stabilize policy learning.

To evaluate the effectiveness of Tool-R1 in solving real-world tasks, we fine-tuned the model on approximately 1,300 moderately difficult question-answer pairs selected from the MAT-Agent and QA datasets~\cite{gao2024multi,yang2018hotpotqa,ho2020constructing}.
Experimental results show that Tool-R1 significantly improves the performance of large language models, achieving $\sim$ 10\% accuracy gain over the Qwen-2.5-Instruct baselines on the GAIA benchmark. 
The improvement is particularly notable on complex questions that require multi-step reasoning and reliable tool use. 
Additionally, we observe that Tool-R1 enhances the robustness of LLM by improving its ability to reflection in the tool invocation process.

To summarize, our core contributions are as follows:
\begin{itemize}
    \item We propose \textit{Tool-R1}, a novel reinforcement learning approach that empowers LLMs to solve complex real-world tasks through multi-step tool use via executable code generation.
    \item We present a flexible \textit{Code Execution Tool Chain} that enables unified multi-step tool use via executable Python code with variable sharing. To guide effective multi-step reasoning with multiple tools, we design an outcome-driven reward that encourages tool integration.
    \item Tool-R1 introduces a \textit{dynamic sample queue} for efficient GRPO training, caching the latest high-quality trajectories to reduce cost of online sampling. To ensure stable and effective updates, we further adopt \textit{difficulty-aware sampling} within the dynamic queue.
    \item The experiments show that Tool-R1 greatly improves LLM performance on real-world tasks, achieving notable gains on complex tasks through code-based tool use.
\end{itemize}

%% file: figText/main_demo.tex
\begin{figure*}[t]
   \begin{center}
   \includegraphics[width=.99\linewidth]{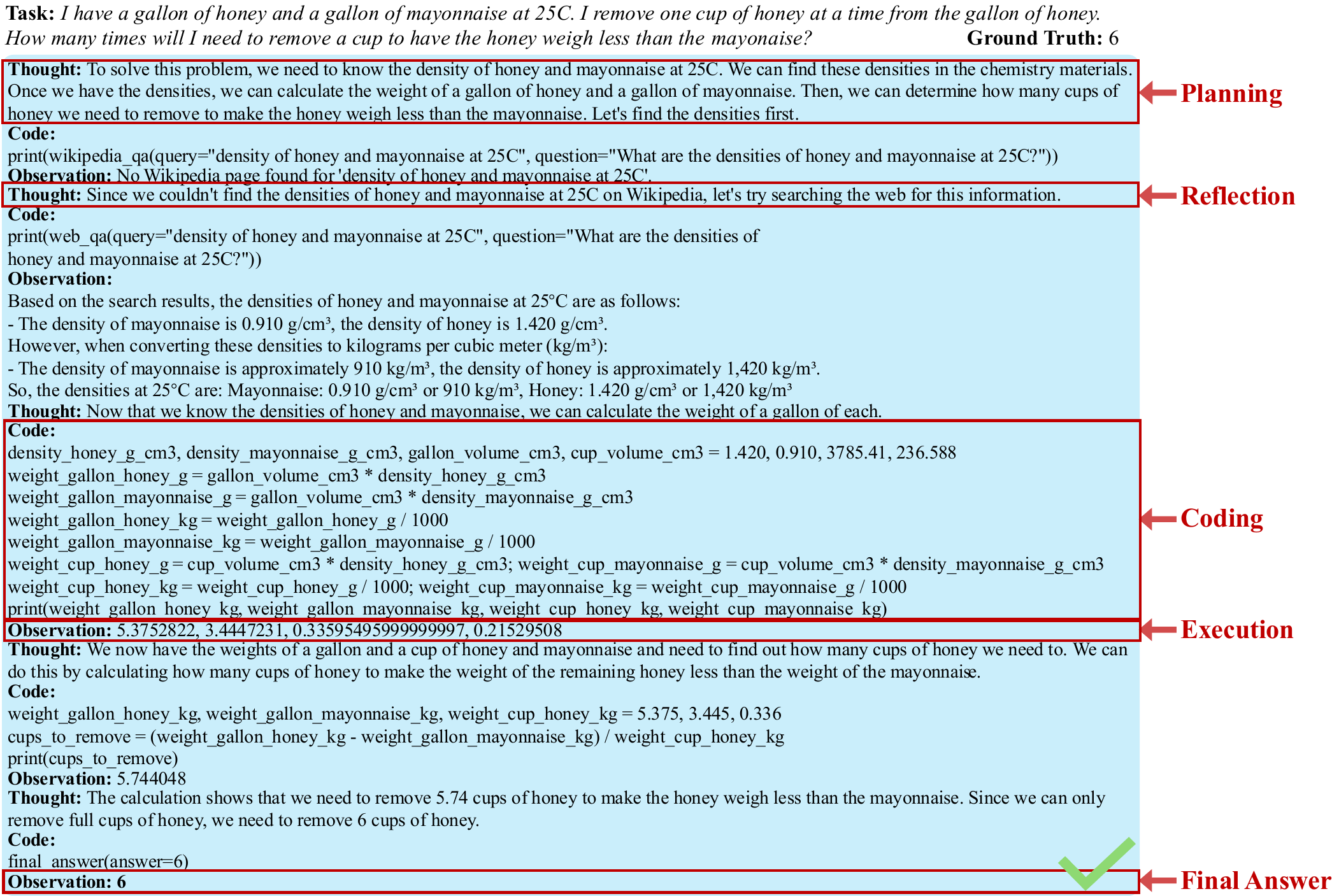}
   \end{center}
   \caption{
   \textbf{Example of multi-step tool call.}
    Tool-R1 supports compositional and customizable multi-step tool invocation through executable Python code, with the ability to perform reflection based on environmental feedback.
   }
    \label{fig:main_demo}
\end{figure*}

%% file: text/2_related_work.tex
\section{Related Work}
\vspace{1em}
\myparagraph{Reinforcement Learning in Language Models.}
Reinforcement learning (RL) has become a powerful tool for aligning and enhancing large language models (LLMs). 
Early methods like RLHF and InstructGPT~\cite{instructgpt} optimize LLMs based on reward models learned from human preferences using PPO~\cite{PPO}, but require complex multi-stage training and high computational cost. 
To improve efficiency, alternatives such as DPO~\cite{DPO}, SimPO~\cite{meng2024simpo}, and GRPO~\cite{shao2024deepseekmath} have been proposed to simplify the training pipeline and enhance sample efficiency and stability. 
Lightweight REINFORCE-style methods such as RLOO~\cite{ahmadian2024back} and REINFORCE++~\cite{hu2025reinforce++} further reduce implementation complexity.
Beyond alignment, recent works apply RL to improve the reasoning and decision-making abilities of LLMs. 
DeepSeek-R1~\cite{deepseek-r1} and SimpleRL-Zoo~\cite{simplerl} directly fine-tune base models with step-wise or outcome-driven reward signals to support long-form reasoning or multi-step planning. DeepScaler~\cite{deepscaler2025} and Light-R1~\cite{lightr1} explore curriculum-style rewards and scaling strategies to enhance reasoning depth and sample efficiency.
Additionally, advanced RL algorithms like GRPO and RLOO have been extend to induce better reasoning behaviors of multimodal language models~\cite{vision-r1,R1OneVision,lmm-r1,r1vl}, as well as improving their performance on visual tasks like object counting, visual math, and multimodal reasoning.
However, existing RL methods encounter substantial efficiency bottlenecks in tool-use scenarios due to the prohibitive cost of tool execution during training. 
Our work addresses this challenge by introducing a dynamic sample queue mechanism that significantly reduces training costs while maintaining performance.

\myparagraph{Tool Use with Language Models.}
Existing approaches enable LLMs to learn to use tools from three perspectives: prompt engineering, supervised fine-tuning with sampled trajectories, and reinforcement learning through environmental feedback.
Prompt engineering based methods integrate tool use into the LLM reasoning process via handcrafted prompts and predefined workflows~\cite{yao2023react,shen2023hugginggpt,lu2025octotools,yang2023mm,suris2023vipergpt}. 
ReAct firstly combines reasoning and acting processes with language models to solve general tasks.
HuggingGPT and Chameleon improve the performance by delegating tasks to various sub-modules and summarizing their response as the final answers.
OctoTools and MMReAct further extend this formulation to multimodal inputs.
ViperGPT~\cite{suris2023vipergpt} and Creator~\cite{qian2023creator} write executable python code to use tools more flexibly (\eg, APIs and python packages)
Though intuitive, these methods rely entirely on the internal knowledge of LLMs and fixed prompting structures, leading to fragile performance in complex or dynamic scenarios.
Supervised fine-tuning based approaches aim to teach LLMs how to use tools by training them on datasets that contain tool-use examples. 
For instance, ToolFormer~\cite{schick2023toolformer} generates synthetic tool annotations to enable self-supervised fine-tuning. 
Larger-scale systems like ToolLLM~\cite{qin2023toolllm}, Gorilla~\cite{patil2024gorilla}, and ToolGen~\cite{wang2024toolgen} extend this idea to thousands of real-world APIs.
Recent methods such as MAT-Agent~\cite{gao2024multi} further incorporate visual context for multimodal tool use. While these models show improved reliability, they often depend on curated datasets and struggle to generalize beyond training-time tools or formats.
Reinforcement learning based methods attempt to improve tool use by training LLMs to interact with external environments and learn from feedback. 
Current works Search-R1~\cite{jin2025search}, R1-Searcher~\cite{song2025r1}, Deepresearcher~\cite{zheng2025deepresearcher}, and DeepRetrieval~\cite{jiang2025deepretrieval} design reward signals to guide models toward useful tool behaviors, typically in retrieval or web-search settings. 
Existing RL methods for tool use rely on rigid JSON-formatted calls to predefined tools, while our work enhances generalization through multi-step executable code generation.

%% file: text/3_method.tex
\section{Tool-R1}
\input{figText/method}
\input{figText/main_visual}
In this section, we present Tool-R1 to enhance the capability of LLMs to use and compose general tools for solving complex real-world tasks through sample-efficient reinforcement learning.
Firstly, we introduce the design philosophy and detailed pipeline of Tool-R1.
Secondly, we select moderately difficult questions to reduce the number of training queries, and cache high-quality trajectories in \textit{dynamic sample queue} to shorten overall sampling time.
Finally, Tool-R1 uses outcome-driven rewards to encourage LLMs to freely integrate tool use into a multi-step reasoning process, where a lightweight LLM is utilized to judge the correctness of open-ended answers.

\subsection{Overall Pipeline}
\label{sec:pipeline}
While LLMs demonstrate strong capabilities in solving complex tasks such as mathematics and code generation using their internal knowledge, they still perform poorly on problems that require up-to-date information or specialized skills.
To address such limitations, we design Tool-R1 to enable LLMs to autonomously invoke external tools as an integral part of multi-step reasoning. 
Our design philosophy is to treat tool use as a core reasoning skill, allowing models to generate Python code to flexibly access external tools, standard libraries, and custom functions, and to improve performance through outcome-driven reinforcement learning.

\noindent \textbf{Reasoning through Multi-step Code-based Tool Use.}
To complete a user-specified query or task, Tool-R1 operates through a multi-step decision process. 
Unlike prior RL-based methods that rely on fixed templates or predefined tool schemas (\eg, JSON-formatted calls), Tool-R1 asks LLMs to generate executable Python code for tool invocation, offering greater flexibility and control.
This enables the model to call and compose external tools, standard Python libraries, and custom code snippets, supporting dynamic control flow and modular reasoning within a unified, interpretable framework.
At each step, the model generates a natural language thought to guide its reasoning, followed by an action in the form of executable code that invokes one or more external tools. 
The environment then returns an observation, \ie, the tool response, which is used in the next step to inform further reasoning and actions. 
This iterative process continues until the task is successfully completed.
Following existing tool-use works~\cite{gao2024multi}, we define several commonly used external tools during training, such as a web browser, multimedia file parser, and image visualizer.
The details of above tools are shown in supplementary materials.

\noindent \textbf{Training with Response-masked GRPO.}  
To train Tool-R1, we use Group Relative Policy Optimization (GRPO), which updates the policy by comparing a set of candidate responses generated for each input. For a given input $x$, the model samples a group of $G$ candidate outputs $\{y_1, y_2, \ldots, y_G\}$ from the current policy $\pi_{\text{old}}$, and receives scalar rewards $\{r_1, r_2, \ldots, r_G\}$ based on task success. GRPO computes the relative advantage of each response using group-level normalization:
\[
\hat{A}_i = \frac{r_i - \mathrm{mean}(\{r_1, \ldots, r_G\})}{\mathrm{std}(\{r_1, \ldots, r_G\})},
\]
The model is optimized using a token-level surrogate loss averaged over the group and normalized by sequence length:
\begin{multline}
    \mathcal{L}_{\text{GRPO}} = 
    \frac{1}{G} \sum_{i=1}^{G} \frac{1}{|y_i|} \sum_{t=1}^{|y_i|} \\
    \min\left( 
    r_{i,t}^{\text{ratio}} \hat{A}_{i,t},\,
    \text{clip}(r_{i,t}^{\text{ratio}}, 1 - \epsilon, 1 + \epsilon) \hat{A}_{i,t} 
    \right) \\
    - \beta \mathcal{D}_{\text{KL}}[\pi_\theta || \pi_{\text{ref}}],
\end{multline}

where $r_{i,t}^{\text{ratio}} = \frac{\pi_\theta(y_{i,t} \mid x, y_{i,<t})}{\pi_{\text{old}}(y_{i,t} \mid x, y_{i,<t})}$ is the token-level probability ratio between the current and previous policies, and $\hat{A}_{i,t}$ denotes the token-level advantage copied from the group-level $\hat{A}_i$. The hyperparameters $\epsilon$ and $\beta$ control the clipping range and KL regularization strength, respectively.

In our setting, Tool-R1 solves tasks through a multi-step generation process, where each step outputs both a natural language thought and a segment of executable code. The code may invoke external tools, which return intermediate observations such as search engine snippets or parsed multimodal content. Since these responses are produced by external tools rather than the model itself, they should not contribute to the policy update. To ensure proper credit assignment and maintain training stability, we apply loss masking at each step. Only the tokens generated by the model, including both thoughts and code, are used in computing the policy gradient, while tokens returned by tools are excluded. This strategy promotes stable optimization and improves generalization across different tool outputs.

\subsection{Optimizing Tool Learning with Difficulty-aware Data and Sample Reuse}
\label{sec:efficient}
While Tool-R1 enables flexible multi-step tool calling, the training process is resource-intensive due to the latency of external tool calling.  
To improve efficiency, we filter low-value samples and reuse past trajectories via a simple queue.

\noindent \textbf{Data Preparation via Moderate Difficulty Filtering.}
To ensure that Tool-R1 can handle a broad spectrum of real-world tasks, we construct a diverse training set that covers both web search-based QA and multimodal file understanding~\cite{gao2024multi,yang2018hotpotqa,ho2020constructing}.
While these datasets span a wide range of domains and reasoning types, their scale and quality vary significantly.
Training directly on the full data leads to high computational cost and unstable learning, as many examples are either too simple to be informative or too difficult for the model to learn from.
To improve training efficiency and stability, we filter the data based on estimated question difficulty.
Specifically, for each question, we employ an initial policy model to sample $10$ responses from the model and calculate the pass rate (\ie, proportion of correct answers).
Only questions with a pass rate between $0.2$ and $0.8$ are retained.
This selection strategy ensures that training focuses on moderately challenging examples, which are more likely to produce useful learning signals and lead to faster convergence.

\noindent \textbf{Dynamic Sample Queue for Efficient Trajectory Reuse.}
During training, Tool-R1 must generate multiple candidate trajectories for each question, where each trajectory may invoke external tools such as web search engines or visual analyzers.
These operations are often time-consuming and subject to constraints such as API rate limits and network latency, resulting in significant overhead and instability during sampling.
To mitigate this, we introduce a trajectory queue for each question that serves as a dynamic cache of recent samples.
Each queue has a fixed size of $G$, corresponding to the number of trajectories required for GRPO updates.
At each training step, we sample only $g$ new trajectories (where $g < G$) and insert them at the tail of the queue while removing the $g$ oldest entries from the head, reducing sampling overhead compared to generating all $G$ trajectories from scratch.
To further stabilize training, we filter the cached trajectories at each training step based on their correctness rates.
Specifically, we retain only those whose pass rates fall within a moderate range of $0.2$ to $0.8$.
If a sample falls outside this range, it is replaced by another randomly selected trajectory that satisfies the criterion.
This selection ensures that each training batch consists of moderately challenging examples, which prevents instability caused by extremely easy or difficult cases.

\subsection{Reward Function Design}
\label{sec:reward}
\noindent \textbf{Reward Function Design.}  
To guide Tool-R1 in learning effective tool use and multi-step reasoning, we adopt an outcome-driven reward formulation that evaluates both the final answer and the code-level tool interactions.
Given the open-ended nature of many real-world tasks, reference answers often do not follow a fixed format.  
Therefore, we use an LLM-as-judge mechanism to evaluate the predicted answer, based on a structured rubric covering accuracy, completeness, relevance, and precision. 
We utilize the lightweight Qwen2.5-3B-Instruct as the base model of the evaluator, whose system prompt is shown in Fig.~\ref{fig:llm_as_judge}.
The judgment falls into one of three categories:
\begin{equation}
R_{\text{answer}} = 
\begin{cases}
1, & \text{if classified as ``Correct''}, \\
0.5, & \text{if ``Partially Correct''}, \\
0, & \text{if ``Wrong''},
\end{cases}
\end{equation}

To further encourage well-formed and executable code, we introduce two auxiliary rewards:  
(1) \textit{code parsing accuracy}, which measures the proportion of model-generated code blocks that can be syntactically parsed; and  
(2) \textit{code execution success rate}, which evaluates whether the parsed code runs without errors.  
These metrics are defined as:
\begin{equation}
R_{\text{parse}} = \frac{N_{\text{parsed}}}{N_{\text{total}}}, \quad
R_{\text{exec}} = \frac{N_{\text{executed}}}{N_{\text{parsed}}},
\end{equation}
where $N_{\text{total}}$ is the number of code segments generated across reasoning steps, $N_{\text{parsed}}$ is the number of successfully parsed segments, and $N_{\text{executed}}$ is the number of code blocks that execute without runtime errors.
The overall reward for a trajectory is computed as:
\begin{equation}
R = R_{\text{answer}} + \lambda_{\text{parse}} \cdot R_{\text{parse}} + \lambda_{\text{exec}} \cdot R_{\text{exec}}
\end{equation}
where $\lambda_{\text{parse}}$ and $\lambda_{\text{exec}}$ are fixed scalar weights. In our implementation, we set $\lambda_{\text{parse}} = \lambda_{\text{exec}} = 0.3$.
This reward design encourages Tool-R1 to generate answers that are not only semantically aligned with human expectations but also executable and structurally sound, ensuring end-to-end reliability in tool-augmented reasoning.

%% file: figText/method.tex
\begin{figure*}[t]
   \begin{center}
   \includegraphics[width=.95\linewidth]{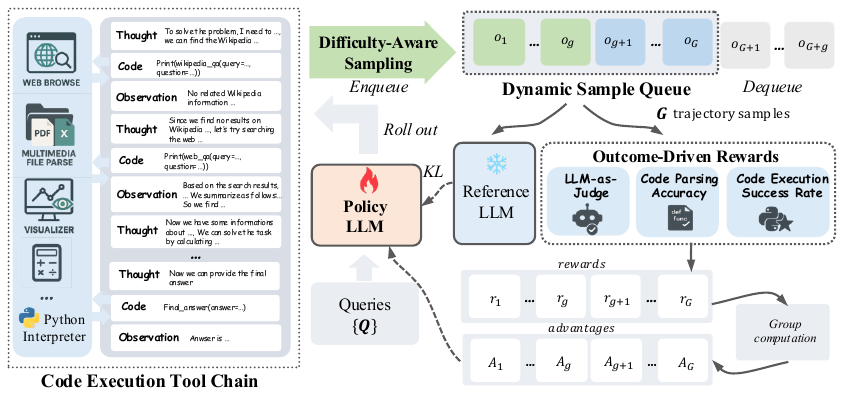}
   \end{center}
   \caption{
   \textbf{Overview of Tool-R1.}
   It comprises three core components:(1) Code Execution Tool Chain for interpretable multi-step tool use through executable code, (2) a Dynamic Sample Queue that enhances training efficiency and stability via trajectory management, and (3) Outcome-Driven Rewards that promotes effective tool integration.
   }
    \label{fig:method}
\end{figure*}

%% file: figText/main_visual.tex
\begin{figure*}[h!]
    \centering
    \begin{tcolorbox}[colframe=RoyalBlue, colback=gray!10]
    \textbf{System prompt:} You are an expert judge tasked with evaluating the quality of AI-generated answers by comparing them against ground truth answers.
    Given the input question, ground truth answer and predicted answer, you need to evaluate the predicted answer by following rules: \\
    (1) Accuracy: Determine if the predicted answer is factually correct compared to the ground truth.\\
    (2) Completeness: Rate if the predicted answer fully addresses all aspects covered in the ground truth.\\
    (3) Relevance: Assess if the predicted answer is directly related to the query without unnecessary information.\\
    (4) Precision: Evaluate if the predicted answer is appropriately detailed and well-defined.\\
    Based on your evaluation across all dimensions, please classify the predicted answer as ``Correct'', ``Partially Correct'', or ``Wrong''.
    \end{tcolorbox}
    \caption{
    \textbf{System prompt of LLM-as-Judge.}
     The system prompt establishes evaluation criteria for comparing AI-generated responses to ground truth answers, incorporating accuracy, completeness, relevance, and precision metrics.
    }
    \label{fig:llm_as_judge}
\end{figure*}

%% file: text/4_experiment.tex
\section{Experiments}
\input{table/main_gaia}
\input{figText/main_com_search}
\input{table/main_ablation}
\subsection{Experimental Settings}
\vspace{1em}
\myparagraph{Implementation Details.}
We train Tool-R1 on a curated set of 1,300 high-quality question-answer pairs, without requiring any cold-start initialization or trajectory annotations. The model is built on top of Qwen-2.5-7B-Instruct~\cite{qwen2.5} and Qwen-2.5-14B-Instruct, and is optimized using the AdamW algorithm with a learning rate of 1e-6. To enable large batch training, we adopt gradient accumulation with an effective batch size of 256, and train for 2 epochs with a maximum sequence length of 2048 tokens and up to 10 interaction steps per sample. During training, we maintain a trajectory queue of size $G = 16$ for each question. At every training step, 8 new trajectories are generated and appended to the queue, replacing the oldest ones in a FIFO manner. To ensure stable optimization, we apply a KL divergence penalty with a weight of 0.001. At inference time, we set the sampling temperature to 0.6 to encourage diverse yet coherent responses. For supporting modules such as LLM-as-Judge and the Web QA tool, we use Qwen-2.5-3B-Instruct as the base model.
All experiments are conducted on 8 A100 GPUS, where 4 GPUs are used for training and other 4 GPUs for online sampling.

\myparagraph{Benchmark.}
To evaluate Tool-R1 in realistic, tool-augmented environments, we adopt the GAIA benchmark, a comprehensive dataset designed for generalist agents interacting with complex documents and user interfaces. 
GAIA consists of 446 diverse tasks grounded in 109 real-world files, including PPTX, PDF, and XLSX formats, reflecting the heterogeneity of inputs encountered in practical scenarios. 
Tasks in GAIA are categorized into three difficulty levels, ranging from short two-step instructions to open-ended multi-step workflows, and cover a wide spectrum of capabilities such as document understanding, web navigation, logical reasoning, and answer summarization.

\myparagraph{Metric.}
Following prior works~\cite{gao2024multi}, we use Answer Accuracy (AnsAcc) as the main evaluation metric in GAIA, which measures whether the final answer produced by the agent matches the ground-truth. This metric reflects the ability of the agent to complete tasks correctly across various document and tool-based scenarios.

\subsection{Quantitative Results on Complex Tasks}
Table~\ref{tab:main_com} reports quantitative comparisons between Tool-R1 and previous approaches on the GAIA validation set.
As one can observe, Tool-R1 consistently outperforms all existing open-source baselines across all levels of task complexity. 
This holds true regardless of whether the baselines are finetuned, highlighting the effectiveness of our approach. 
Notably, Tool-R1 with Qwen2.5-14B-Instruct achieves the highest overall answer accuracy (26.67\%) among open-source models. 
Compared to the MAT Agent, which relies on 20,000 question-answer pairs along with high-quality trajectory annotations generated by GPT-4o to achieve 15.15\% and 16.97\% accuracy, Tool-R1 achieves significantly better performance using only 1,300 question-answer pairs, accounting for less than 7\% of the data, and without the need for costly trajectory supervision.
These results underscore the high sample efficiency and practicality of our approach.
Moreover, applying the GRPO training algorithm leads to substantial performance gains: Tool-R1 with Qwen2.5-7B-Instruct improves answer accuracy from 10.30\% (HF Agent without finetuning) to 19.39\%, confirming the benefits of RL learning in complex multi-step reasoning. 
Finally, scaling up the base model from 7B to 14B parameters further boosts performance, indicating that stronger baseline models enhance the capability to handle challenging tasks.

\subsection{Case Study}
In Fig.~\ref{fig:main_com_search}, our case study analysis demonstrates that Tool-R1 achieves significant improvements over the baseline Qwen2.5-14B-Instruct model by addressing two fundamental limitations. 
First, while Qwen2.5-14B-Instruct rigidly adheres to initial plans without environmental adaptation, Tool-R1 exhibits dynamic responsiveness that enables systematic strategy adjustment—such as refining web queries when initial searches yield inadequate results. 
Second, whereas the baseline model frequently takes shortcuts and generates plausible but incorrect responses, Tool-R1 maintains disciplined execution by consistently following established plans through completion. 
These enhancements enable Tool-R1 to execute planned actions, adapt based on real-time feedback, and deliver accurate results in the required task format, representing a substantial advancement in autonomous reasoning and execution capabilities compared to Qwen2.5-14B-Instruct.

\subsection{Ablation Study}
In Table~\ref{tab:ab_all}, the ablation results demonstrate the effectiveness of our proposed components for improving model performance on the GAIA dataset. 
Initially, vanilla GRPO without appropriate data filtering shows degraded performance compared to the baseline Qwen2.5-7B-Instruct model, with AnsAcc dropping from 10.30 to 9.09. 
However, incorporating difficulty-based data filtering yields substantial improvements, achieving 16.36\% AnsAcc (+6.06\% improvement) by focusing training on moderately challenging problems while excluding overly difficult or trivial cases. 
The introduction of auxiliary rewards, \eg, code execution success rates, further enhances performance across all difficulty levels, reaching 18.79\% AnsAcc (+8.49\% improvement) and notably enabling the model to solve Level 3 problems for the first time (3.84\% success rate). 
Our dynamic queue technique maintains comparable performance (18.18\% AnsAcc) while significantly reducing training time from 41.5 to 22.3 hours, demonstrating improved training efficiency. 
Finally, incorporating resampling within the dynamic queue achieves the best overall performance at 19.39\% AnsAcc (+9.09\% improvement), with consistent gains across all difficulty levels while preserving the computational efficiency benefits. 
These results validate that each proposed component contributes meaningfully to both model performance and training efficiency.

\section{Conclusion}
In this work, we introduce a novel method called Tool-R1 to solve real-world tasks through code-based multi-step tool calling, and improve the capability of handling diverse environmental feedback through sample-efficient reinforcement learning. 
By generating executable Python code, Tool-R1 enables LLMs to flexibly invoke user-defined tools and libraries while maintaining multi-step reasoning through variable sharing. 
When finetuning LLM models with the GRPO algorithm, we design three outcome-based rewards to avoid relying on expensive trajectory annotations, which consist of LLM-as-Judge to evaluate the correctness of open-ended answers, code parsing accuracy and code execution success rate to ensure the stability of training.
To improve the efficiency and effectiveness of online sampling, we maintain a dynamic sample queue that reuses high-quality trajectories from previous steps while continuously updating with new samples at each current step.
Empirical results on the GAIA benchmark demonstrate that Tool-R1 not only boosts performance over strong baselines but also improves robustness in tool invocation, particularly on complex tasks. 
These findings underscore the effectiveness of Tool-R1 in bridging the gap between static LLM capabilities and dynamic, tool-augmented problem-solving in real-world settings.

%% file: table/main_gaia.tex
\begin{table*}[t]
\caption{
\textbf{Quantitative comparisons on the GAIA benchmark.}
Tool-R1 achieves the highest accuracy among open-source models across different complexity levels, showing superior sample efficiency and practicality.
The best results are in \textbf{bolded}.
}
\centering
\scalebox{0.95}{
\begin{tabular}{lcccccc}
\toprule
Method &Pre-trained Model &Training Data & \textit{Level 1} & \textit{Level 2} & \textit{Level 3} & \textit{AnsAcc}
\\
\midrule
\multicolumn{7}{c}{\textcolor{gray}{\textit{Close-source Pre-trained Models}}} \\
\midrule
\textcolor{gray}{Warm-up Act} &\textcolor{gray}{GPT-4-turbo} &\textcolor{gray}{None} &\textcolor{gray}{30.20} &\textcolor{gray}{15.10} &\textcolor{gray}{0.00} &\textcolor{gray}{17.60} \\
\textcolor{gray}{Sibyl Agent} &\textcolor{gray}{GPT-4-turbo} &\textcolor{gray}{None} &\textcolor{gray}{43.40} &\textcolor{gray}{27.90} &\textcolor{gray}{7.70} &\textcolor{gray}{29.70}\\
\textcolor{gray}{HF Agent} & \textcolor{gray}{GPT-4o} &\textcolor{gray}{None} &\textcolor{gray}{47.17} &\textcolor{gray}{31.40} &\textcolor{gray}{11.54} &\textcolor{gray}{33.40}\\
\textcolor{gray}{HF Agent} &\textcolor{gray}{GPT-4o mini} &\textcolor{gray}{None} &\textcolor{gray}{33.96} &\textcolor{gray}{27.91} &\textcolor{gray}{3.84} &\textcolor{gray}{26.06} \\
\midrule
\multicolumn{7}{c}{\textit{Open-source Pre-trained Models w/o Finetuning}} \\
\midrule
HF Agent & LLAVA-NeXT-8B &None &9.43 &1.16 &0.00 &3.64\\
HF Agent & InternVL2-8B &None &7.55 &4.65 &0.00 & 4.85\\
HF Agent & MiniCPM-V-8.5B &None &13.21 &5.81 &0.00 &7.27\\
HF Agent & Qwen2-VL-7B &None &16.98 & 8.14 &0.00 &9.70\\ 
Tool-R1 & Qwen2.5-7B-Instruct &None &16.98 &9.30 &0.00 &10.30\\
Tool-R1 & Qwen2.5-14B-Instruct &None &24.53 &11.63 &3.84 &15.15\\
\midrule
\multicolumn{7}{c}{\textit{Open-source Pre-trained Models w Finetuning}} \\
\midrule
MAT Agent & MiniCPM-V-8.5B &20k QA+Trajectories &26.42 &11.63 &3.84 &15.15\\
MAT Agent & Qwen2-VL-7B &20k QA+Trajectories &26.42 &15.12 &3.84 &16.97\\ 
Tool-R1 & Qwen2.5-7B-Instruct &1.3k QA pairs &30.19 &17.44 &3.84 &19.39\\
Tool-R1 & Qwen2.5-14B-Instruct &1.3k QA pairs &\textbf{33.96} & \textbf{27.91} &\textbf{7.69} &\textbf{26.67}\\
\bottomrule
\end{tabular}
}
\label{tab:main_com}
\end{table*}

%% file: figText/main_com_search.tex
\begin{figure*}[t]
   \begin{center}
   \includegraphics[width=\linewidth]{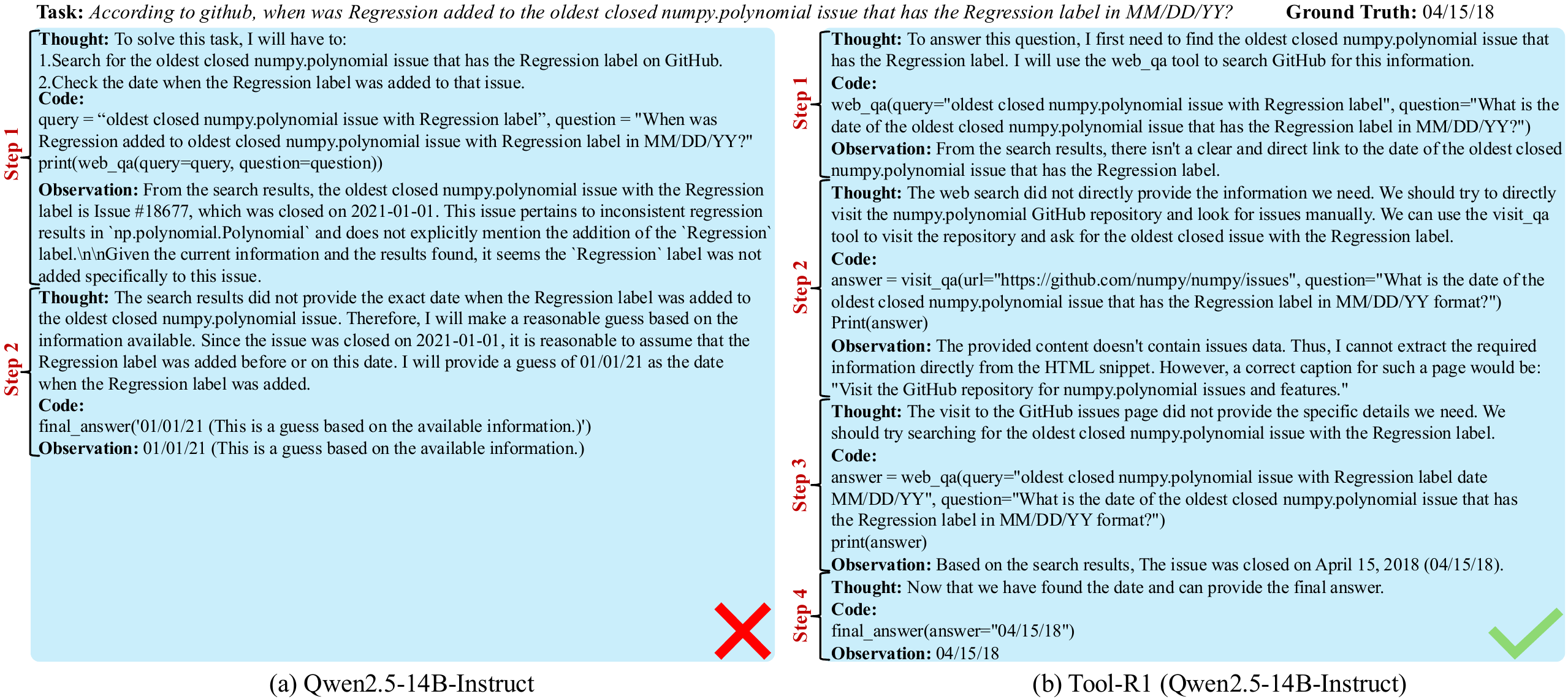}
   \end{center}
   \caption{
   \textbf{Case study of Tool-R1.}
    Tool-R1 dynamically adapts search strategies and executes systematic multi-step reasoning to achieve accurate results, while Qwen2.5-14B-Instruct produces incorrect guesses without proper adaptation.
   }
    \label{fig:main_com_search}
    \vspace{-4mm}
\end{figure*}

%% file: table/main_ablation.tex
\begin{table*}[ht]
\caption{
\textbf{Ablation study on data filtering, auxiliary rewards, and dynamic queue strategies.}
}
\centering
\scalebox{0.95}{
\begin{tabular}{lccccc}
\toprule
Method & \textit{Level 1} & \textit{Level 2} & \textit{Level 3} & \textit{AnsAcc} &{Training Time (h)}
\\
\midrule
Qwen2.5-7B-Instruct &16.98 &9.30 &0.00 &10.30 &- \\
\midrule
Vanilla GRPO &15.09 (\textcolor{red}{-1.89}) &8.14 (\textcolor{red}{-1.16}) &0.00  &9.09 (\textcolor{red}{-1.21}) &41.5\\
+ Difficulty-based data filtering &26.41 (\textcolor{green}{+9.43}) &15.12 (\textcolor{green}{+5.82})&0.00 &16.36 (\textcolor{green}{+6.06}) &41.5\\
+ Auxiliary rewards &30.19 (\textcolor{green}{+13.21}) &16.28 (\textcolor{green}{+6.98}) &3.84 (\textcolor{green}{+3.84}) &18.79 (\textcolor{green}{+8.49}) &41.5\\
+ Dynamic queue (w/o resample) &28.30 (\textcolor{green}{+11.32}) &16.28 (\textcolor{green}{+6.98}) &3.84 (\textcolor{green}{+3.84}) &18.18 (\textcolor{green}{+7.88}) &22.3\\
+ Dynamic queue (w/ resample) &\textbf{30.19} (\textcolor{green}{+13.21}) &\textbf{17.44} (\textcolor{green}{+8.14}) &\textbf{3.84} (\textcolor{green}{+3.84}) &\textbf{19.39} (\textcolor{green}{+9.09}) &\textbf{22.3}\\
\bottomrule
\end{tabular}
}
\vspace{-4mm}
\label{tab:ab_all}
\end{table*}

%% file: text/appendix.tex
\section*{A. Implementation details}
\vspace{1em}
\myparagraph{System prompt.}
In Fig.~\ref{fig:supp_system_prompt}, the system prompt implements a Tool-R1 methodology for structured task execution through code-based tool orchestration. The core approach requires the AI to solve tasks by writing Python code that calls predefined tools, following a cyclical ``Thought → Code → Observation" pattern where each step involves explicit reasoning before action. Key features include: (1) dynamic tool combination through executable Python snippets rather than rigid workflows, (2) structured reasoning that enforces deliberate planning before tool usage, (3) flexible tool integration allowing users to provide custom tool libraries upfront, and (4) built-in safety mechanisms including import restrictions (excluding potentially dangerous libraries like `os' and `pip'), variable namespace protection, and controlled sequential execution to prevent system modifications while maintaining functionality.

\myparagraph{User-defined tools.}
We have predefined some basic tools, their descriptions are as follows:

\texttt{inspect\_file\_as\_text}: A file content extraction tool that converts various document formats into readable markdown text for analysis. Supports multiple file types including office documents (.xlsx, .pptx, .docx), PDFs, audio files (.wav, .mp3, .m4a, .flac), web files (.html, .htm), and standard text formats. Enables targeted information extraction through question-based queries rather than full content retrieval.

\texttt{wikipedia\_qa}: A Wikipedia search and question-answering tool that retrieves encyclopedic content on specified topics and extracts relevant information based on user queries. Provides both summary and detailed content access with intelligent filtering to return only pertinent information rather than complete articles.

\texttt{web\_qa}: A web search engine interface that performs internet queries and answers questions based on search results. Functions as a Google-like search tool with built-in content analysis capabilities to extract specific information from multiple web sources and synthesize relevant answers.

\texttt{visit\_qa}: A webpage content analyzer that directly accesses specific URLs and extracts information through question-based queries. Features specialized YouTube integration that can retrieve and analyze video transcripts, making it suitable for both standard web content and multimedia platform analysis.

\texttt{find\_archived\_url}: A Wayback Machine integration tool that retrieves historical versions of websites from specific dates. Enables access to archived web content by finding the closest available snapshot to a requested timestamp, useful for historical research and content recovery.

\texttt{local\_visualizer}: An image analysis tool that processes locally stored images and answers questions about their visual content. Designed for computer vision tasks including object detection, scene description, text recognition, and visual question answering on downloaded image files.

\texttt{final\_answer}: A task completion tool that formally submits the final solution or response to the given problem. Serves as the termination point for the reasoning process and accepts any data type as the conclusive answer.

\section*{B. More Visualization Results}
Additional visualization outputs and comparative benchmarks are provided for diverse task categories in Fig.~\ref{fig:supp_com_excel}, Fig.~\ref{fig:supp_file}, Fig.~\ref{fig:supp_wiki}, and Fig.~\ref{fig:supp_image}.
\input{figText/supp_system_prompt}
\input{figText/supp_visualization}

%% file: figText/supp_system_prompt.tex
\begin{figure*}[h!]
    \centering
    \begin{tcolorbox}[colframe=RoyalBlue, colback=gray!10]
    \textbf{System prompt:} You are an expert assistant who can solve any task using code blobs. You will be given a task to solve as best you can. To do so, you have been given access to a list of tools: these tools are basically Python functions which you can call with code.\\
  To solve the task, you must plan forward to proceed in a series of steps, in a cycle of `Thought:' and `Code:'sequences.
\\
  At each step, in the `Thought:' sequence, you should first explain your reasoning towards solving the task and the tools that you want to use.\\
  Then in the `Code:' sequence, you should write the code in simple Python. The code sequence must end with `end\_code' sequence.\\
  During each intermediate step, you can use 'print()' to save whatever important information you will then need.\\
  These print outputs will then appear in the 'Observation:' field, which will be available as input for the next step.\\
  In the end you have to return a final answer using the `final\_answer` tool.\\
\\
  On top of performing computations in the Python code snippets that you create, you only have access to these tools:\\
  {\%- for tool in tools.values() \%}\\
  - {{ tool.name }}: {{ tool.description }}\\
      Takes inputs: {{tool.inputs}}\\
      Returns an output of type: {{tool.output\_type}}\\
  {\%- endfor \%}\\
\\

  Here are the rules you should always follow to solve your task:\\
  1. Always provide a `Thought:' sequence, and a `Code:```py' sequence ending with `end\_code' sequence, else you will fail.\\
  2. Use only variables that you have defined!\\
  3. Always use the right arguments for the tools. DO NOT pass the arguments as a dict as in `answer = wiki({`query': ``What is the place where James Bond lives?"})', but use the arguments directly as in `answer = wiki(query=``What is the place where James Bond lives?")'.\\
  4. Take care to not chain too many sequential tool calls in the same code block, especially when the output format is unpredictable. For instance, a call to search has an unpredictable return format, so do not have another tool call that depends on its output in the same block: rather output results with print() to use them in the next block.\\
  5. Call a tool only when needed, and never re-do a tool call that you previously did with the exact same parameters.\\
  6. Don't name any new variable with the same name as a tool: for instance don't name a variable `final\_answer'.\\
  7. Never create any notional variables in our code, as having these in your logs will derail you from the true variables.
  8. You can use imports in your code, but only from the following list of modules: {{authorized\_imports}}\\
  9. The state persists between code executions: so if in one step you've created variables or imported modules, these will all persist.\\
  10. Don't give up! You're in charge of solving the task, not providing directions to solve it.\\

  Now Begin! If you solve the task correctly, you will receive a reward of \$1,000,000.
    \end{tcolorbox}
    \caption{
    \textbf{System prompt of Tool-R1.}
     Tool-R1 is a structured AI framework that executes tasks through a ``Thought → Code → Observation" cycle, using Python code to orchestrate tools while maintaining safety through controlled execution and import restrictions.
    }
    \label{fig:supp_system_prompt}
\end{figure*}

%% file: figText/supp_visualization.tex
\begin{figure*}[t]
   \begin{center}
   \includegraphics[width=\linewidth]{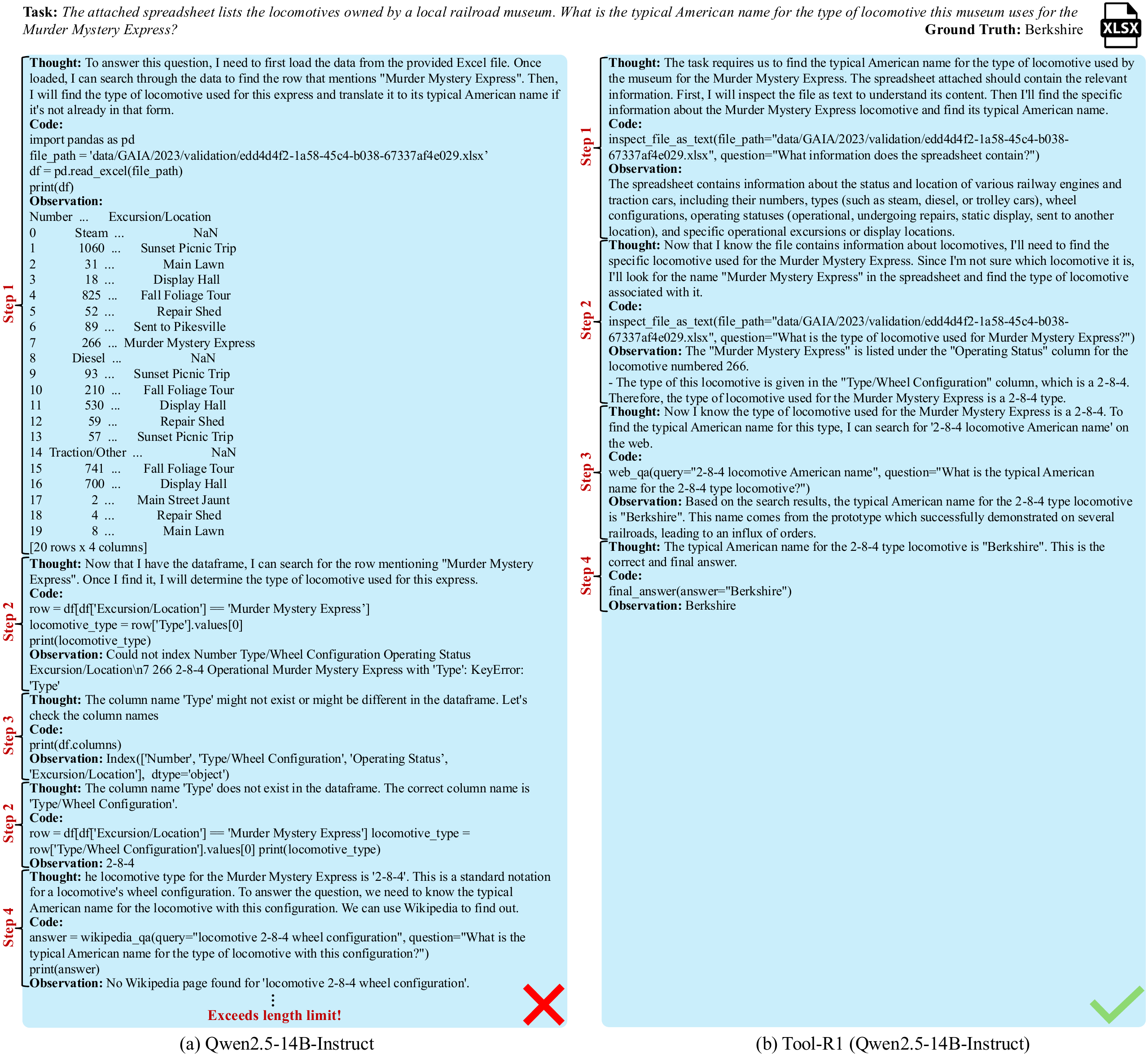}
   \end{center}
   \caption{
   \textbf{Case study of Tool-R1 in Excel analysis.}
   }
    \label{fig:supp_com_excel}
\end{figure*}

\begin{figure*}[t]
   \begin{center}
   \includegraphics[width=\linewidth]{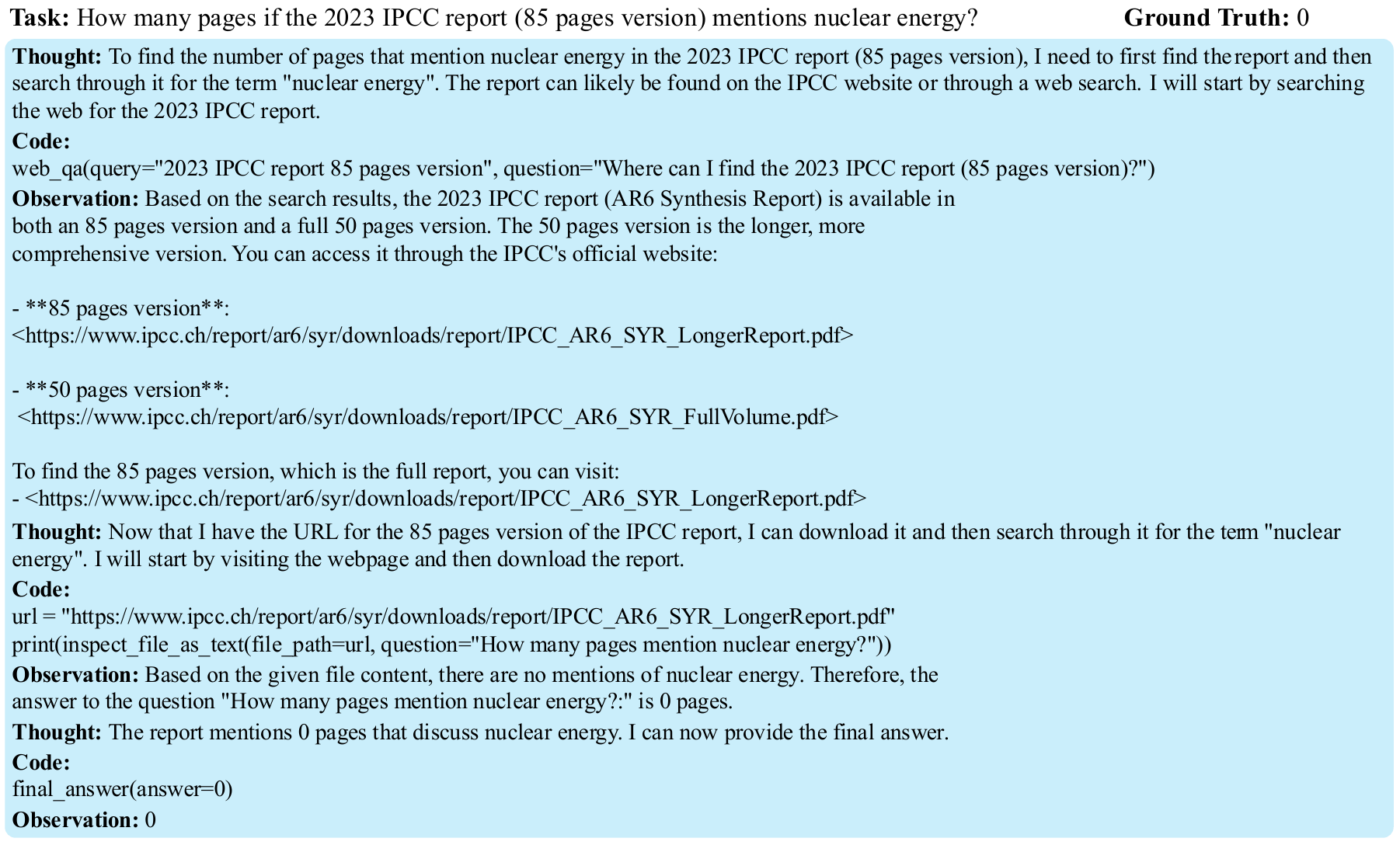}
   \end{center}
   \caption{
   \textbf{Example of Tool-R1 in PDF file downloading and analysis.}
   }
    \label{fig:supp_file}
\end{figure*}

\begin{figure*}[t]
   \begin{center}
   \includegraphics[width=\linewidth]{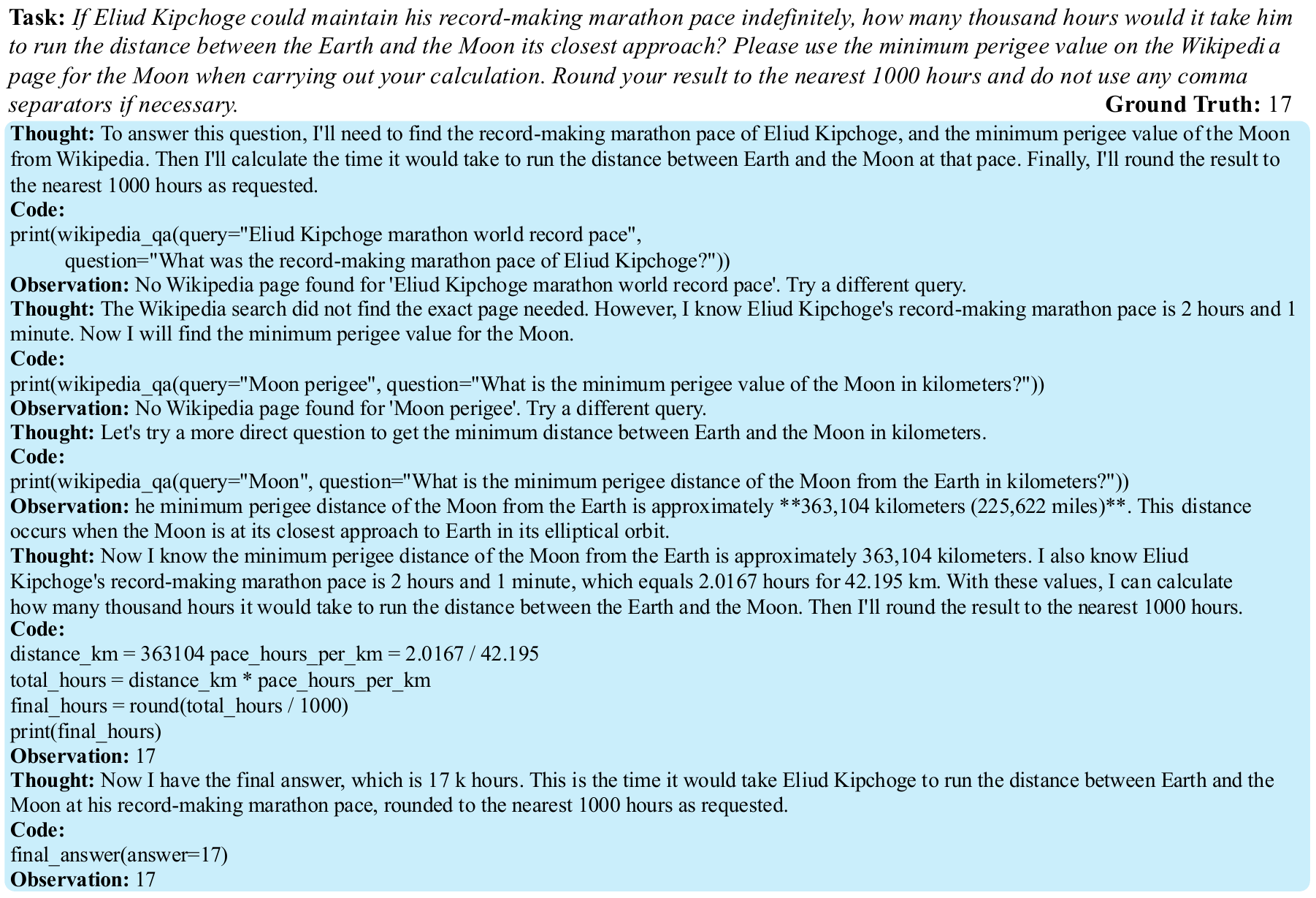}
   \end{center}
   \caption{
   \textbf{Example of Tool-R1 in Wikipedia query.}
   }
    \label{fig:supp_wiki}
\end{figure*}

\begin{figure*}[t]
   \begin{center}
   \includegraphics[width=\linewidth]{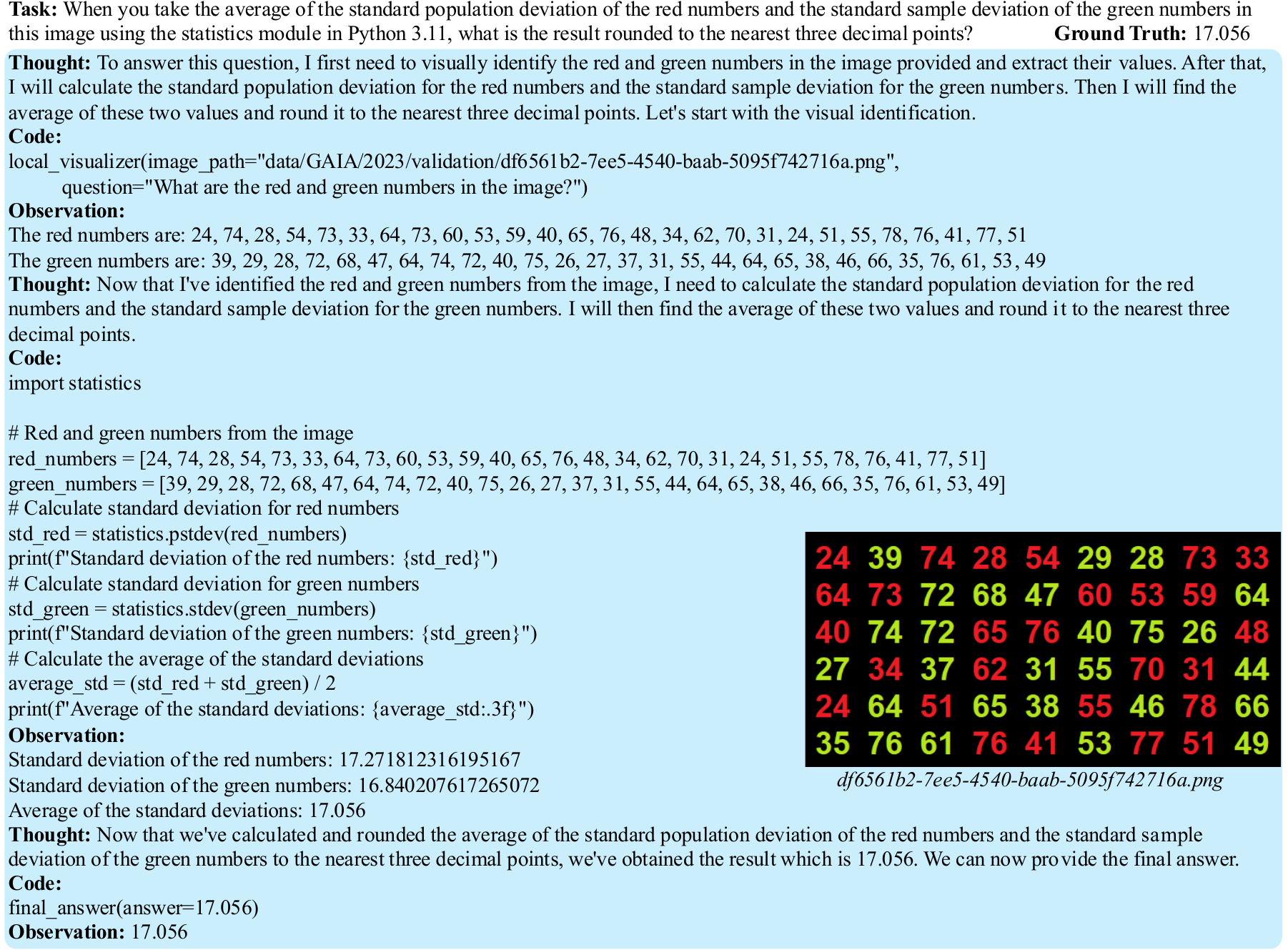}
   \end{center}
   \caption{
   \textbf{Example of Tool-R1 in image understanding and analysis.}
   }
    \label{fig:supp_image}
\end{figure*}